\documentclass{article}
\usepackage{arxiv}
\usepackage[bookmarks=false]{hyperref}

\hypersetup{nolinks=true}

\usepackage[utf8]{inputenc} % allow utf-8 input
\usepackage[T1]{fontenc}    % use 8-bit T1 fonts

\usepackage{booktabs}       % professional-quality tables
\usepackage{amsfonts}       % blackboard math symbols
\usepackage{nicefrac}       % compact symbols for 1/2, etc.
\usepackage{microtype}      % microtypography
\usepackage{lipsum}		% Can be removed after putting your text content
\usepackage{graphicx}
\usepackage{doi}
\usepackage[numbers]{natbib}

\usepackage{amsmath}

\DeclareMathOperator*{\argmin}{arg\,min}

\usepackage{chngcntr}
\counterwithin*{section}{part}
\usepackage{graphicx}

\title{\emph{TLGAN}: document Text Localization using Generative Adversarial Nets}

\author{
	\And
	\And
	\And
	\And
	Dongyoung Kim\thanks{Corresponding author, http://www.dykim.net} \\
	Data Analytic Laboratory\\
	Samsung Life Insurance\\
	Seoul, South Korea \\
	\texttt{dongyoung.kim@me.com} \\
	\And
	\And
	\And
	\And
	Myungsung Kwak\\
	Data Analytic Laboratory\\
	Samsung Life Insurance\\
	Seoul, South Korea \\
	\texttt{yesmung@gmail.com} \\
	\And
	\And
	\And
	\And
	\And
	Eunji Won\\
	Data Analytic Laboratory\\
	Samsung Life Insurance\\
	Seoul, South Korea \\
	\texttt{weji1216@gmail.com} \\
	\And
	Sejung Shin\\
	Data Analytic Laboratory\\
	Samsung Life Insurance\\
	Seoul, South Korea \\
	\texttt{way3shin@gmail.com} \\
	\And
	Jeongyeon Nam\\
	Data Analytic Laboratory\\
	Samsung Life Insurance\\
	Seoul, South Korea \\
	\texttt{ckleckle@gmail.com} \\
}

\renewcommand{\shorttitle}{\emph{TLGAN}: document Text Localization using Generative Adversarial Nets}

\hypersetup{
pdftitle={TLGAN: document Text Localization using Generative Adversarial Nets},
pdfsubject={cs:ml},
pdfauthor={Dongyoung Kim},
pdfkeywords={Deep learning, Generative adversarial network, Image processing, optical character recognition, computer vision},
}

\begin{document}
\maketitle

\begin{abstract}
Text localization from the digital image is the first step for the optical character recognition task. Conventional image processing based text localization performs adequately for specific examples. Yet, a general text localization are only archived by recent deep-learning based modalities. Here we present document Text Localization Generative Adversarial Nets (TLGAN) which are deep neural networks to perform the text localization from digital image. TLGAN is an versatile and easy-train text localization model requiring a small amount of data. Training only ten labeled receipt images from Robust Reading Challenge on Scanned Receipts OCR and Information Extraction (SROIE), TLGAN achieved $99.83\%$ precision and $99.64\%$ recall for SROIE test data. Our TLGAN is a practical text localization solution requiring minimal effort for data labeling and model training and producing a state-of-art performance.
\end{abstract}

% keywords can be removed
%%%%%%%%%%%%%%%%%%%%%%%%%%%%%%%%%%%%%%%%%%%%%
%NOTE: THE VERSION HAS CITATION PROBLEM AND WE WILL FIX IT SOON.
\section{Introduction}
% requirements of text localization from document images
In enterprise business, printed documents are a major communication tool. These papers are often acquired using optical devices like scanners or cameras and the data are compressed/stored as digital images. Such document images contain valuable information and there is a big need to make digital images to interpretable text. Optical character recognition (OCR) is a method to translate the printed document to digital text. OCR processes are to localize text in images following by text recognition at loci \citep{Tafti2016}. Further text/language processes may be added as needs. The text localization task is to detect texts from digital images which may contain not only texts but also graphics, drawings, lines, and noises. Conventional image processing techniques can be applied and works for specific examples, yet, such approaches are vulnerable to the real-world noises which may not be described or may not be able to describe at the processing algorithm.

% text localization studies using deep learning
Recent advances in deep learning show a great success in object detection. There are two major approaches for the object detection: region proposal network (RPN) and semantic segmentation. RPN searches object boundary coordinates, i.e. regions of interest (ROIs), and is successfully demonstrated by faster region proposal CNN (Faster-RCNN) \citep{Ren2015}, single shot multibox detector (SSD) \citep{Liu2015}, you only look once (YOLO) \citep{Redmon2015} with their successors \citep{Dai2016,Redmon2016,Lin2016,Fu2017,Redmon2018}. The RPN based object detection modalities has been fine-tuned for text detection tasks, e.g. TextBoxes \citep{Liao2016}, fully-convolutional regression network (FCRN) \citep{Gupta2016}, efficient and accurate scene text detector (EAST) \citep{Zhou2017} and more. Semantic segmentation produces segmentation map corresponding to the object locations and shapes. Fully convolutional networks (FCN) \citep{Long2014}, U-Net \citep{Ronneberger2015} with following approaches \citep{Yu2015, Drozdzal2016, Jegou2016, Kim2019} et cetera are examples. Semantic segmentation approaches has been adopted and modified for the text detection and certainly has become a powerful text detection tool. Semantic segmentation based text detectors such as character region awareness for text detection (CRAFT) \citep{Baek2019a}, multi oriented corner text detectors \citep{Lyu2018}, pixel aggregation network (PANNet) \citep{Wang2019}, and connectionist text proposal network (CTPN) \citep{Tian2016} are top ranked at robust reading competition of focused scene text (FST) and scanned receipt OCR (SROIE) \citep{Karatzas2013,ICDARRRC2019,ICDAR2019}. Certainly, PANNet and CTPN used ImageNet pretrained VGG network \citep{Deng2009,Simonyan2014} for feature extraction from text contained images and the both models show great performances \citep{Karatzas2013,ICDARRRC2019,ICDAR2019}.

% gan for semantic segmentation
Generative adversarial network (GAN) is a framework to train a deep learning model using an adversarial processes \citep{Goodfellow2014}. Several GAN models have devised and, especially, GAN shows brilliant results for image-to-image translation problem, e.g. creating semantic segmentation from image or the reverse, drawing to photo translation, enhancing image resolution and more \citep{Luc2016, Isola2017, Ledig2016, Park2019}. Recent studies found GAN for the object detection from images and show superior and versatile object detection performance \citep{Prakash2019, Liu2019, Wang2020}.

% TLGAN, motivatil, techinical breif, rank results, easy train results
Here we introduce a document text localization generative adversarial network (TLGAN), which is a GAN specially designed for detecting text location in document images. TLGAN follows the semantic segmentation based text localization approach \citep{Baek2019a, Lyu2018, Wang2019, Tian2016} and estimates precise text location using a generator network structuring in a set of residual convolutional layers  \citep{Ledig2016,Johnson2016,Ioffe2015,He2015,Shi2016}. Effective text location estimation is carried out using Imagenet pretrained VGG network \citep{Deng2009,Simonyan2014} and TLGAN uses VGG as discriminator loss evaluation function rather than feature extraction unit for semantic segmentation \citep{Ledig2016,Wang2019,Tian2016}. TLGAN, therefore, take benefits of VGG's great feature extractions without having large VGG computation in addition to versatile performance of an adversarial learning process \citep{Prakash2019, Liu2019, Wang2020}. TLGAN achieved $99.83\%$ precision and $99.64\%$ recall for reading challenge on Scanned Receipts OCR and Information Extraction (SROIE). Notably, we found TLGAN learned text location with a samll set of data, i.e. ten labeled images are enough to reproduce similar performance at SROIE dataset. Our TLGAN is a practical text localization solution requiring minimal effort for data labeling and model training and producing a state-of-art performance.

% contribution of the paper

%%%%%%%%%%%%%%%%%%%%%%%%%%%%%%%%%%%%%%%%%%%%%

\section{Methods}\label{sec:methods}
\begin{figure}[htp]
	\centering
	\includegraphics[scale=0.7]{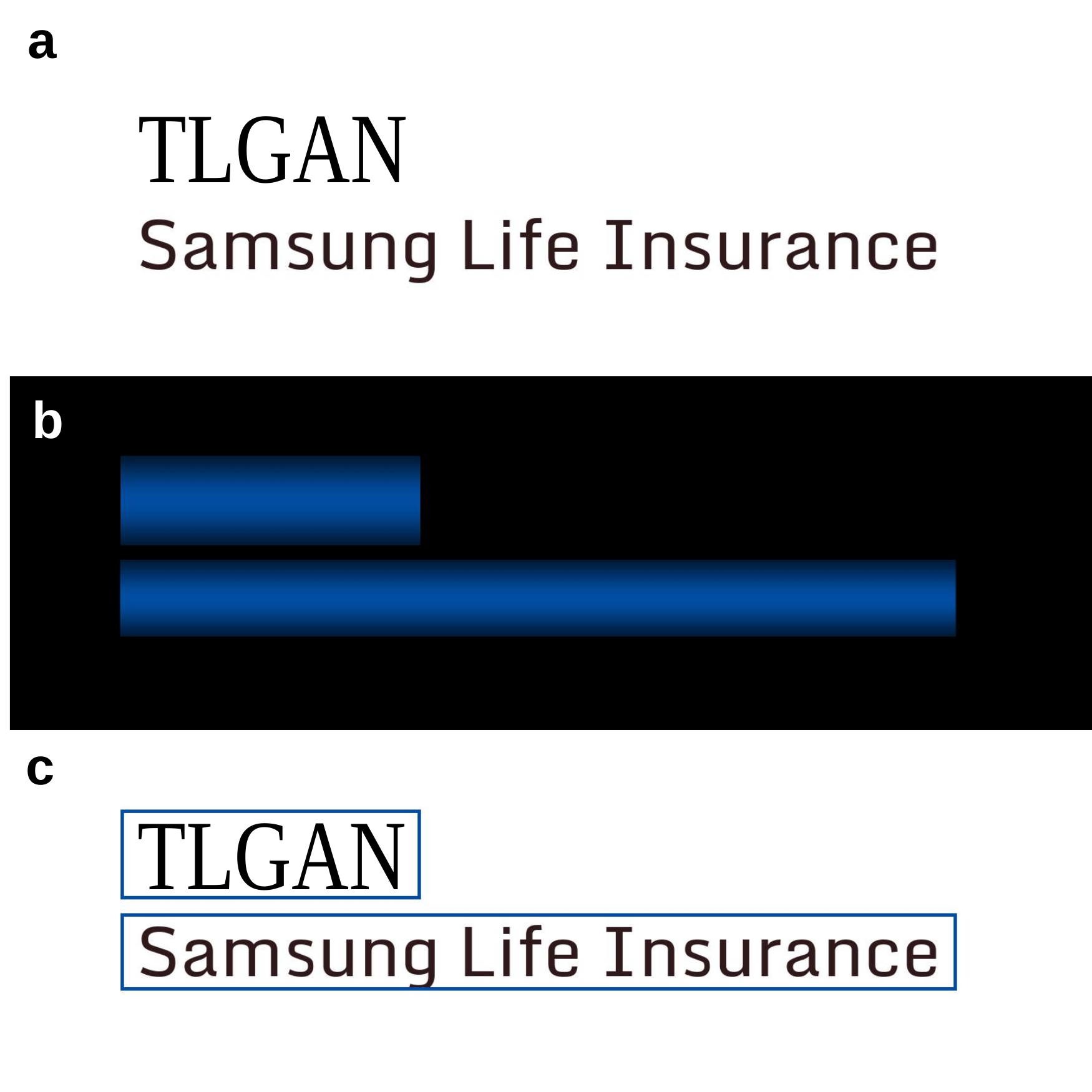}
	\caption{An example of (a) text image and corresponding (b) text localization map (colored in blue) and (c) text localization result (blue boxes).}
	\label{fig:textmapandbox}
\end{figure}

Document Text Localization Generative Adversarial Nets (TLGAN) aims to perform text localization from a text-containing image $I^{R}$ (Fig. \ref{fig:textmapandbox}a) by estimating a text localization map $I^{M}$ (Fig. \ref{fig:textmapandbox}b). Here we first address the formation of text localization map $I^{M}$ (section \ref{sec:methods:textlocalizationmap}) and TLGAN architecture (section \ref{sec:methods:TLGAN}) followed by text localization approach (section \ref{sec:methods:textlocalization} and Fig. \ref{fig:textmapandbox}c). We next show the training and evaluation strategies of the model used in the manuscript in sections \ref{sec:methods:training} and \ref{sec:methods:training}.

%--------------------------------------------

\subsection{Text localization map}\label{sec:methods:textlocalizationmap}
Let we have an image $I^{R}$ containing $n$ text positioning at $\vec{P}_n, n=0,...,N$. A cylindrical Gaussian map image $\vec{M}$ is given:
\begin{equation}
M(mx, my) = \frac{1}{2\pi\sigma_{my}} e^{-my^2/2\sigma_{my}^2},
\end{equation}
where $\sigma_{my}$ denote the width of the cylindrical Gaussian map, $(mx, my) \in \mathbb{R}^2$.

The text locations are marked in $I^M$ by wrapping a cylindrical Gaussian maps $\vec{M}$ into each text position $\vec{P}_n$ into $I^{M}$ using a set of affine transformations \citep{Baek2019a},
\begin{equation}
I^M = \sum^N_{n=1}\vec{P}_n = \sum^N_{n=1} \left( A_n\cdot\vec{M} + \vec{b}_n\right),\qquad n=0,1,2,...,N.
\end{equation}

Figure \ref{fig:textmapandbox}b shows an example text localization map corresponding to the text image in figure \ref{fig:textmapandbox}a.

%--------------------------------------------

\subsection{TLGAN}\label{sec:methods:TLGAN}
A convolutional neural network (CNN) $G_{p_g}$ parameterized by $p_g$ is devised to estimate a text localization map $I^M$ from a text-containing image $I^R$:
\begin{equation}
G_{p_g}:I^R \mapsto I^M,
\end{equation}
where $p_g$ denotes a set of weights and biases in deep neural nets. We find the $\hat{p}_g$ by solving the equation \ref{eq:generator} over $K$ training images $I^R$ and the corresponding maps $I^M$:
\begin{equation}\label{eq:generator}
\hat{p}_g = \argmin_{p_g} \frac{1}{K} \sum^K_{k=1} l^M (G_{p_g}(I^R), I^M).
\end{equation}

From equation \ref{eq:generator}, $l^M$ is a loss function defined:
\begin{equation}\label{eq:loss}
l^M = \frac{1}{J}\sum_j^J \left( q\times(I^M_j-G_{p_g}(I^R)_j)^2 + r\times(\phi(I^M)_j-\phi(G_{p_g}(I^R))_j)^2 \right), \quad j=0,1,2,...,J
\end{equation}
where $j$ denotes $J$ pixels in the image, $q$ and $r$ denote the weights of the loss contents, $\phi$ is a feature extraction function which is a inter-layer feature output from a pretrained CNN, e.g. VGG19 \citep{Simonyan2014} ImageNet pretrained \citep{Ledig2016,Dong2014,Shi2016,Gatys2015,Bruna2015}.

Here we utilized a CNN network $D_{p_d}$ parameterized by $p_d$ following generative adversarial nets (GAN) \citep{Goodfellow2014}. Note we are following the successful work of SR-GAN by Ledig et al. \citep{Ledig2016}. Both $G_{p_g}$ and $D_{p_d}$ are CNNs and the detailed architecture of the networks are shown in Supplementary Information \ref{sup:networkarchitecture}. Briefly, $G_{p_g}$ takes input image $I^R$ and consists of stacks of residual blocks which composed by convolutional layers, batch normalization layers, and parametric ReLU activation layers \citep{Johnson2016,Ioffe2015,He2015,Shi2016}. The features from residual blocks are computed using a $s$ strided convolution layer. The output $I^M$ is given by a fianl convolution layer with $\tanh$ activation. $D_{p_d}$ is a convolutional neural network to discriminate $I^M$ to $\hat{I^M}=G_{p_g}(I^R)$ and is composed by a set of convolution blocks with a convolution layer, batch normalization layer and leaky ReLU activation layer \citep{Radford2015}. A final dense layer with the sigmoid activation makes discrimination between $I^M$ to $\hat{I^M}$ using features from convolution blocks.

To find $\hat{G}_{p_g}$, $G_{p_g}$ and $D_{p_d}$ are optimized alternately by solving min-max problem in equation \ref{eq:gan} \citep{Goodfellow2014}:
\begin{equation} \label{eq:gan}
\min_{p_g}\max_{p_d} \mathbb{E}_{I^M}\left(\log D_{p_d}(I^M)\right)+\mathbb{E}_{I^R}\left(\log-D_{p_d}(G_{p_g}(I^R)\right).
\end{equation}

%--------------------------------------------

\subsection{Image preprocessing and post processing} \label{sec:methods:prepostprocessing}
Images were resized and their intensities are adjusted for train and test. We first detect a content region of the image which is the area containing information rather empty space. The content area is computed by summing the pixel intensities over x and y axis and by finding front and back edges of the signal assuming the contents exist in a rectangular region. The images are resized at the certain content region size, i.e. 550 pixels in short axis for the data set used in section \ref{sec:methods:sroie}. Image intensities between its 50\% and 99.95\% of the maximum value were mapped to the value between 0 and 255. The scaled images were processed by inferencing a trained TLGAN model. The inference output is scaled back to the original image size via the bicubic interpolation followed by the text localization described in secion \ref{sec:methods:textlocalization}.

%--------------------------------------------

\subsection{Text localization}\label{sec:methods:textlocalization}
Trained CNN $G_{p_g}$ generates a text localization map $\hat{I^M}$ described in section \ref{sec:methods:textlocalizationmap}. The $\hat{I^M}$ annotates text locations as a set of cylindrical Gaussian maps shown in figure \ref{fig:textmapandbox}b. Image segmentation over $\hat{I^M}$ were performed using a simple threshold followed by morphological image processes of the dilation and the border following method \citep{Suzuki1985,Baek2019a}. Rectangular bounding boxes were found from the segmented images (figure \ref{fig:textmapandbox}c).

%--------------------------------------------
\subsection{Training details and parameters}\label{sec:methods:training}
All the models were trained on machines configured with an Intel Xeon W-2135 CPU and a NVIDIA GTX 1080 Ti GPU. TLGAN model $G_{pg}$ was constructed with a convolution layer stride $s=4$ and loss parameters $q = 1$ and $r = 0.001$ in equation \ref{eq:loss} (see section \ref{sec:methods:TLGAN}). Models were optimzied using the Adam optimizer with learning rate $\alpha=0.0002$, $\beta_1=0.5$, $\beta_2=0.999$ and $\epsilon=10^{-7}$ \citep{Kingma2015}. All the models, training and inference were implemented and tested using Tensorflow (https://www.tensorflow.org/) version 2.4.0 and Python version 3.6.10 with Ubuntu version 18.04.5 LTS.

%%%%%%%%%%%%%%%%%%%%%%%%%%%%%%%%%%%%%%%%%%%%%

\section{Experiments}
%--------------------------------------------
\subsection{Benchmark dataset and model}\label{sec:methods:sroie}
A set of experiments were performed using Robust Reading Challenge on Scanned Receipts OCR and Information Extraction (SROIE) dataset by International Conference on Document Analysis and Recognition (ICDAR) \citep{ICDAR2019}. $626$ and $361$ scanned receipt images with text bounding box annotations were given from ICDAR SROIE for training and test, subsequently. A TLGAN model was trained using the training dataset provided from SROIE. Images in training dataset were randomly cropped (augmented) $600$ times in the size of $128 \times 128$ pixels on its width and height due to the limited graphic memory on our system. The TLGAN model was trained using $62,600$ augmented images. The data were randomly sampled in batch size 8 and the model was trained over $120,000$ mini-batches. The training hyper-parameters were given in section \ref{sec:methods:training}.

Experimental results from TLGAN were evaluated by following SROIE evaluation protocol and SROIE evaluation software \citep{ICDAR2019,Karatzas2013}. Briefly, SROIE evaluation protocol is implemented based on DetEval \citep{Wolf2006}. The SROIE evaluation program computes the mean average precision and the average recall based on F1 score \citep{Everingham2015}. H-mean score is defined by the average of the mean average precision and the average recall. We refer the results of ICDAR SROIE website (https://rrc.cvc.uab.es/?ch=13\&com=evaluation\&task=1) accessed at Oct., 19, 2020 to make the comparison table in table \ref{table:sroierank} \citep{ICDAR2019}.

\begin{figure}[htp]
	\centering
	\includegraphics[scale=1]{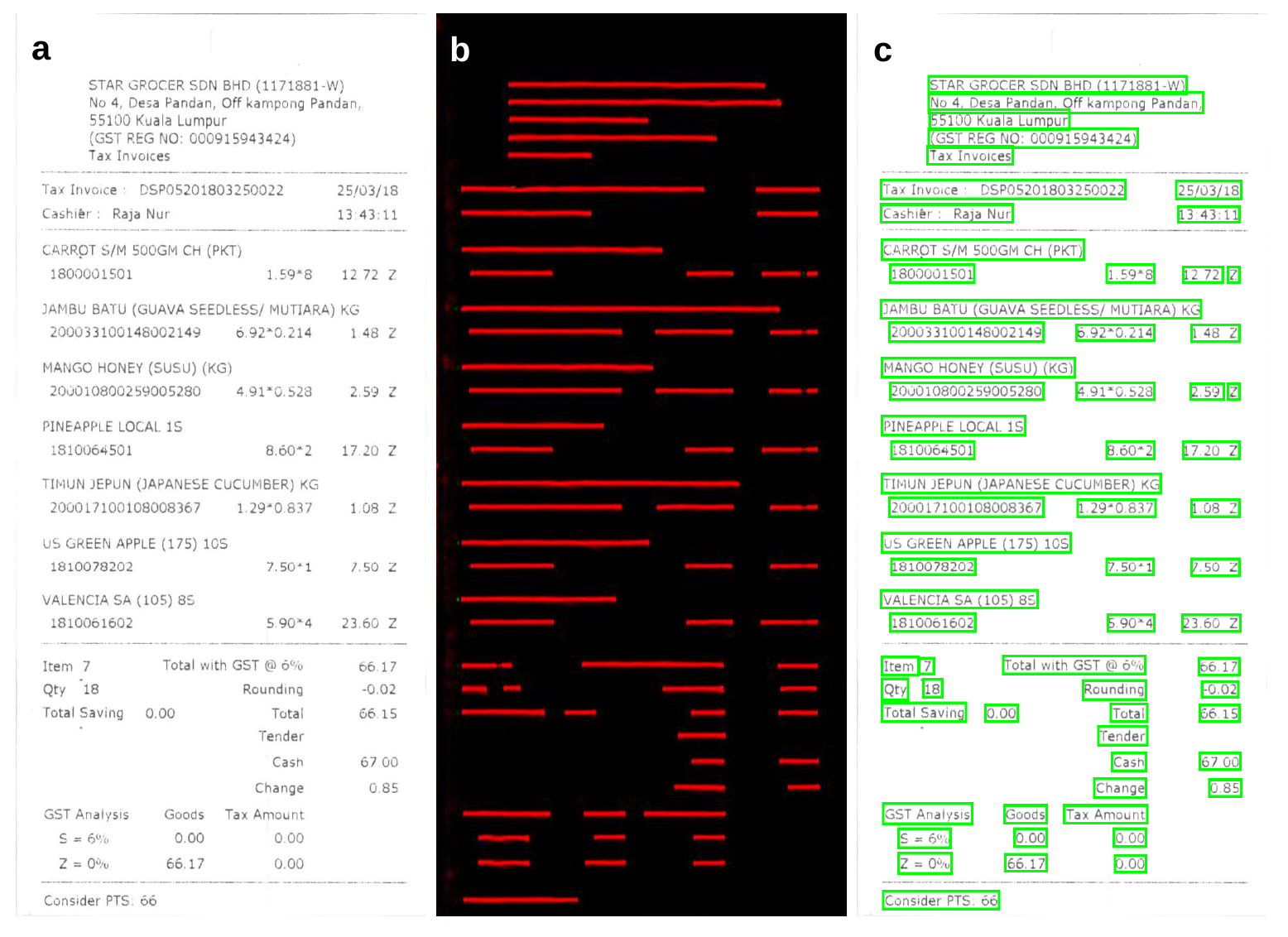}
	\caption{An example of (a) a scanned receipt image from SROIE and corresponding (b) text localization map generated using a TLGAN model. (c) text localization result presented in green boxes.}
	\label{fig:example_sroie}
\end{figure}

\subsection{Results}\label{sec:experiments:result}
Figure \ref{fig:example_sroie} shows an example of text localization of a SROIE data predicted using a TLGAN model. A TLGAN model generated a text localization map (figure \ref{fig:example_sroie}b) from a preprocessed image of figure \ref{fig:example_sroie}a, and the text locations were identified via localization process (figure \ref{fig:example_sroie}c). The TLGAN model was tested using the SROIE task1 test dataset and the model outperformed by achieving $99.83\%$ precision, $99.64\%$ recall, and $99.23\%$ hmean (see table \ref{table:sroierank}). The TLGAN consists of 1.45 million parameters (see supplementary table \ref{sup:networkarchitecture}) which is much smaller than image classification networks such as a VGG16 (138M, \citep{Simonyan2014}) and a ResNet (25M, \citep{He2015}). Besides its performances and sizes, we found the TLGAN models were trained easily that the model training is saturated after few thousand epochs. We hypothesised that the TLGAN model learns text localization features only with few labeled images. To verify this, we conducted a set of experiments in section \ref{sec:experiments:expsmall} making TLGAN models with subsets of training data.

%--------------------------------------------
\begin{table}[]
	\caption{\label{table:sroierank}Experimental results of TLGAN and others for SROIE taks 1, 2020-10-19 accessed \citep{ICDARRRC2019}.}
	\centering
	\begin{tabular}{@{}lllllll@{}}
		\toprule
		\textbf{Rank}       & \textbf{Date}                & \textbf{Method}                                  & \textbf{Recall}           & \textbf{Precision}        & \textbf{Hmean}            & \textbf{Ref.}      \\ \midrule
		\textit{\textbf{-}} & \textit{\textbf{2020-10-19}} & \textit{\textbf{TLGAN (ours)}}                   & \textit{\textbf{99.64\%}} & \textit{\textbf{99.83\%}} & \textit{\textbf{99.91\%}} & \textit{\textbf{}} \\
		1                   & 2020-08-10                   & BOE\_AIoT\_CTO                                   & 98.76\%                   & 98.92\%                   & 98.84\%                   &   \citep{Wang2019, Lyu2018} \\
		2                   & 2019-04-22                   & SCUT-DLVC-Lab-Refinement                         & 98.64\%                   & 98.53\%                   & 98.59\%                   &   N.A.  \\
		3                   & 2019-04-22                   & Ping An Property \& Casualty Insurance    & 98.60\%                   & 98.40\%                   & 98.50\%                   &     \citep{Sun2019,Zhou2017,Zhang2019b}  \\
		4                   & 2019-04-22                   & H\&H Lab                                         & 97.93\%                   & 97.95\%                   & 97.94\%                   &    \citep{Lyu2018,Zhou2017}  \\
		5                   & 2020-09-27                   & Only PAN                                         & 96.51\%                   & 96.80\%                   & 96.66\%                   &    \citep{Wang2019} \\
		6                   & 2019-04-22                   & GREAT-OCR Shanghai University                    & 96.62\%                   & 96.21\%                   & 96.42\%                   &    \citep{Gao2019,Wang2019a} \\
		7                   & 2019-04-23                   & BOE\_IOT\_AIBD                                   & 95.95\%                   & 95.99\%                   & 95.97\%                   &    \citep{Redmon2018}\\
		8                   & 2019-04-23                   & EM\_ocr                                          & 95.85\%                   & 96.08\%                   & 95.97\%                   &    N.A.\\
		9                   & 2019-05-10                   & Clova OCR                                        & 96.04\%                   & 95.79\%                   & 95.92\%                   &     N.A.\\
		10                   & 2019-04-21                   & IFLYTEK-textDet\_v3                              & 93.77\%                   & 95.89\%                   & 94.81\%                   &   \citep{Tian2016} \\ \bottomrule
	\end{tabular}
\end{table}

\begin{figure}[htp]
	\centering
	\includegraphics[scale=0.8]{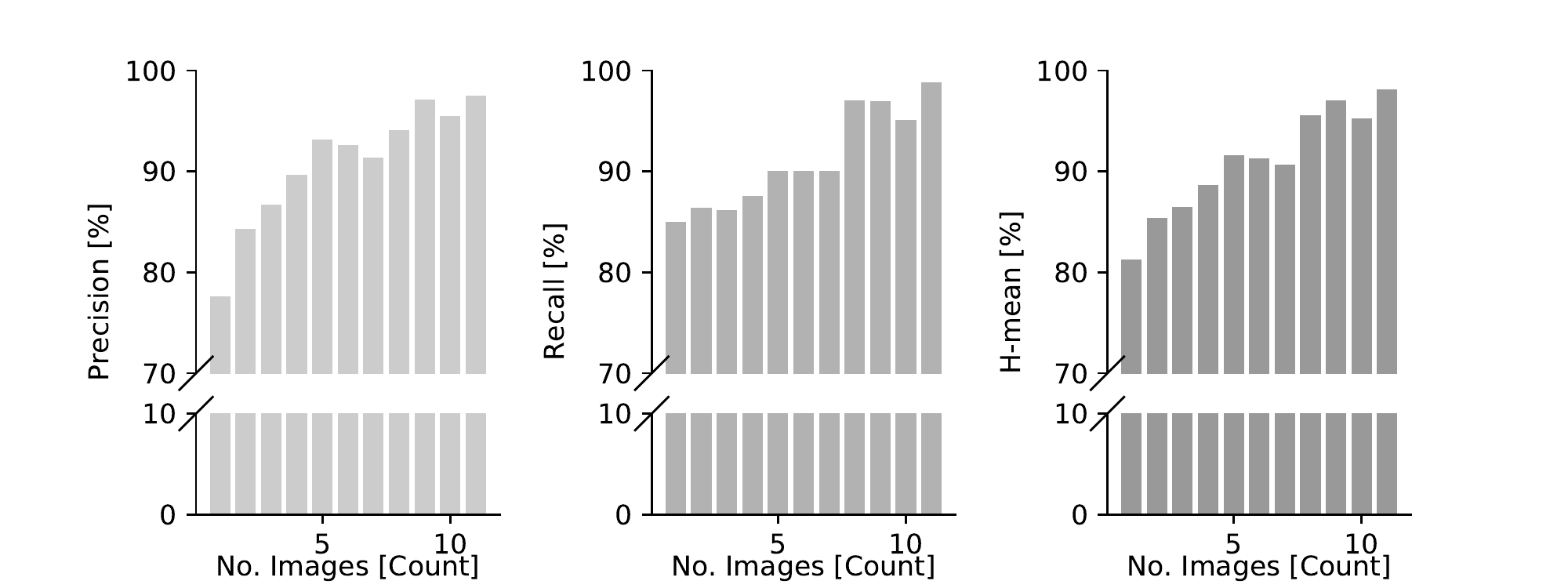}
	\caption{SROIE evaluation results from TLGAN model trained by $n$ labeled images, $n=1,2,...,11$. Precision, recall and H-mean shows from left to right panels, subsequently. }
	\label{fig:sroie_smalldata}
\end{figure}

%--------------------------------------------
\subsection{Experiments with a subset of training data} \label{sec:experiments:expsmall}
A set of TLGAN models were trained using subsets of training data, i.e. eleven images $I_n, n=1,2,...,11$ were randomly sampled from the training dataset of ICDAR SROIE, and ten TLGAN models $G_n, n=1,2,...,11$ were trained by sampling $n$ images followed by preprocessing described in sections \ref{sec:methods:prepostprocessing} and \ref{sec:methods:training}. Figure \ref{fig:sroie_smalldata} shows results from eleven TLGAN models tested using ICDAR SROIE test dataset. The test precision and recall is over $90\%$ hmean at a TLGAN model trained with only five labeled images. Further, the TLGAN model trained with 11 images almost reached to the state-of-art scores ($97\%$ hmean). For document detection tasks, our TLGAN model needs minimal amount of training data significantly reducing the data labeling works.
%--------------------------------------------

%%%%%%%%%%%%%%%%%%%%%%%%%%%%%%%%%%%%%%%%%%%%%

\section{Discussions}
% summary,
TLGAN is a deep learning model to detect text in document and is trained using a generative adversarial network (GAN) approach. A generator network in the TLGAN model finds the text location by translating a scanned document image into a text localization map  followed by finding text bounding boxes from the map (figure \ref{fig:textmapandbox}). To train a generator model, a discriminator network and a feature extraction network forms adversarial losses to find both image contents and features. A TLGAN model was trained for ICDAR SROIE task 1 dataset \citep{ICDARRRC2019} and recorded $99.83\%$ precision, $99.64\%$ recall, and $99.23\%$ hmean (table \ref{table:sroierank}). Further, we found TLGAN learns TLGAN location easily that having ten labeled document images make good text detection model with TLGAN approach. Also, the TLGAN generator network used in the manuscript is defined using 1.45M parameter, which is smaller than many other image processing networks.

% good things
The TLGAN uses image features from a ImageNet pretrained VGG19 network as an adversarial loss. The effective TLGAN feature extraction from a VGG network is successfully demonstrated \citep{Deng2009,Simonyan2014} and the pretrained VGG network is used as a part of their model. Here, we rather uses the large VGG model (e.g. VGG16 with 138M parameters) but indirectly added in to a adversarial loss in addition to the mean-square-error loss. We found the TLGAN rapidly learns text location specific features from a pretrained VGG network  devoiding an expensive computation of a VGG network. Such a knowledge transfer may allow the TLGAN not only to learn the text location map fast but also to need a small set of training data.

% limitation
The current implementation of the TLGAN has following limitations. First, the generator network of TLGAN follows residual convolutional network design \citep{Ledig2016,Johnson2016,Ioffe2015,He2015,Shi2016} and it certainly helps on a stable adversarial learning \citep{Ledig2016}. Yet, a residual convolutions with $3\times3$ kernels only find the convolutional receptive field locally, therefore, the generator is limited to learn the image features at a certain size. In other words, the text contents in the image requires in specific font size and image resolution to achieve the best result with the current TLGAN implementation (see section \ref{sec:methods:prepostprocessing}). We found such can be solved by replacing residual convolutional layers to a U-Net like architecture \citep{Ronneberger2015} which forms multi-resolution features with a high-resolution image reconstruction. We certainly experience some GAN issues of vanishing gradients and mode collapse issues with a U-Net like generator in a long batch training. Here we uses VGG19 to take the feature loss from the image. Second, although this VGG network is not a part of a generator, it is a part of training and is taking a large computation and memory during the training. We tried replacing the VGG19 to the MobileNetV2 \citep{Sandler2018} which reduces the computation and memory uses during the training as well as maintains learning features similar to the VGG19. This certainly a better option in practice to have larger batches at the training. Third, the generator in TLGAN only has 1.45M parameters, yet the model solves pixel to pixel problem and computation is expensive. In addition, the ROI proposal computation remains on CPU computation as a post processing (see section \ref{sec:methods:prepostprocessing}). Such a ROI proposal can be integrated within the network \citep{Ren2015} or implementing post processing module on GPU.

% contribution, conclusion
TLGAN is a document text localization GAN to form a text localization map from the document image followed by the text localization. The TLGAN takes the advantages of the adversarial learning and the pretrained convolutional network of VGG and learns the text localization features rapidly and easily. TLGAN is a practical text localization model by reducing the data labeling work significantly, and can be trained readily for a new kind of datasets. Further investigations are needed to prove the benefits and limitations of the TLGAN.

%%%%%%%%%%%%%%%%%%%%%%%%%%%%%%%%%%%%%%%%%%%%%
% reference

\newpage

\bibliographystyle{IEEEtran}

%\input{manuscript_TLGAN.bbl}
%\input{\jobname.bbl}
%\bibliography{reference}

%%%%%%%%%%%%%%%%%%%%%%%%%%%%%%%%%%%%%%%%%%%%%
\section*{Acknowledgements} elsarticle
We thank all the data analytic laboratory members at Samsung Life Insurance for their support and feedback.

\section*{Author information}
%--------------------------------------------
\subsection*{Contributions}
D.K. and M.K. designed the research; D.K., M.K., E.W., S.S. and J.N. performed the research; D.K. wrote the manuscript. D.K and M.K contribute the manuscript equally.
%--------------------------------------------
\subsection*{Corresponding author}
Correspondence to Dongyoung Kim (dongyoung.kim@me.com; http://www.dykim.net).
%--------------------------------------------
\section*{Ethics declarations}
%--------------------------------------------
\subsection*{Competing interests}
D.K., M.K. and E.W. are inventors of the filed patents in Korea; 10-2019-0059652 (KR).

%%%%%%%%%%%%%%%%%%%%%%%%%%%%%%%%%%%%%%%%%%%%%

\clearpage
\appendix
\section{Supplementary Tables}
\subsection{TLGAN generator network architectures}\label{sup:networkarchitecture}
\begin{table}[htp]
	\centering
	{\tiny
		\begin{tabular}{@{}llll@{}}
			\toprule
			\textbf{Layer (type)}                         & \textbf{Output Shape}       & \textbf{Param \#} & \textbf{Connected to}                                            \\ \midrule
			input\_4 (InputLayer)                         & {[}(N, N, N){]} & 0                 &                                                                  \\
			conv2d\_8 (Conv2D)                            & multiple                    & 15616             & input\_4{[}0{]}{[}0{]}                                           \\
			activation (Activation)                       & multiple                    & 0                 & conv2d\_8{[}0{]}{[}0{]}                                          \\
			conv2d\_9 (Conv2D)                            & multiple                    & 36928             & activation{[}0{]}{[}0{]}                                         \\
			activation\_1 (Activation)                    & multiple                    & 0                 & conv2d\_9{[}0{]}{[}0{]}                                          \\
			batch\_normalization\_7 (BatchNormalization)  & multiple                    & 256               & activation\_1{[}0{]}{[}0{]}                                      \\
			conv2d\_10 (Conv2D)                           & multiple                    & 36928             & batch\_normalization\_7{[}0{]}{[}0{]}                            \\
			batch\_normalization\_8 (BatchNormalization)  & multiple                    & 256               & conv2d\_10{[}0{]}{[}0{]}                                         \\
			add (Add)                                     & multiple                    & 0                 & batch\_normalization\_8{[}0{]}{[}0{]}, activation{[}0{]}{[}0{]}  \\
			conv2d\_11 (Conv2D)                           & multiple                    & 36928             & add{[}0{]}{[}0{]}                                                \\
			activation\_2 (Activation)                    & multiple                    & 0                 & conv2d\_11{[}0{]}{[}0{]}                                         \\
			batch\_normalization\_9 (BatchNormalization)  & multiple                    & 256               & activation\_2{[}0{]}{[}0{]}                                      \\
			conv2d\_12 (Conv2D)                           & multiple                    & 36928             & batch\_normalization\_9{[}0{]}{[}0{]}                            \\
			batch\_normalization\_10 (BatchNormalization) & multiple                    & 256               & conv2d\_12{[}0{]}{[}0{]}                                         \\
			add\_1 (Add)                                  & multiple                    & 0                 & batch\_normalization\_10{[}0{]}{[}0{]}, add{[}0{]}{[}0{]}        \\
			conv2d\_13 (Conv2D)                           & multiple                    & 36928             & add\_1{[}0{]}{[}0{]}                                             \\
			activation\_3 (Activation)                    & multiple                    & 0                 & conv2d\_13{[}0{]}{[}0{]}                                         \\
			batch\_normalization\_11 (BatchNormalization) & multiple                    & 256               & activation\_3{[}0{]}{[}0{]}                                      \\
			conv2d\_14 (Conv2D)                           & multiple                    & 36928             & batch\_normalization\_11{[}0{]}{[}0{]}                           \\
			batch\_normalization\_12 (BatchNormalization) & multiple                    & 256               & conv2d\_14{[}0{]}{[}0{]}                                         \\
			add\_2 (Add)                                  & multiple                    & 0                 & batch\_normalization\_12{[}0{]}{[}0{]}, add\_1{[}0{]}{[}0{]}     \\
			conv2d\_15 (Conv2D)                           & multiple                    & 36928             & add\_2{[}0{]}{[}0{]}                                             \\
			activation\_4 (Activation)                    & multiple                    & 0                 & conv2d\_15{[}0{]}{[}0{]}                                         \\
			batch\_normalization\_13 (BatchNormalization) & multiple                    & 256               & activation\_4{[}0{]}{[}0{]}                                      \\
			conv2d\_16 (Conv2D)                           & multiple                    & 36928             & batch\_normalization\_13{[}0{]}{[}0{]}                           \\
			batch\_normalization\_14 (BatchNormalization) & multiple                    & 256               & conv2d\_16{[}0{]}{[}0{]}                                         \\
			add\_3 (Add)                                  & multiple                    & 0                 & batch\_normalization\_14{[}0{]}{[}0{]}, add\_2{[}0{]}{[}0{]}     \\
			conv2d\_17 (Conv2D)                           & multiple                    & 36928             & add\_3{[}0{]}{[}0{]}                                             \\
			activation\_5 (Activation)                    & multiple                    & 0                 & conv2d\_17{[}0{]}{[}0{]}                                         \\
			batch\_normalization\_15 (BatchNormalization) & multiple                    & 256               & activation\_5{[}0{]}{[}0{]}                                      \\
			conv2d\_18 (Conv2D)                           & multiple                    & 36928             & batch\_normalization\_15{[}0{]}{[}0{]}                           \\
			batch\_normalization\_16 (BatchNormalization) & multiple                    & 256               & conv2d\_18{[}0{]}{[}0{]}                                         \\
			add\_4 (Add)                                  & multiple                    & 0                 & batch\_normalization\_16{[}0{]}{[}0{]}, add\_3{[}0{]}{[}0{]}     \\
			conv2d\_19 (Conv2D)                           & multiple                    & 36928             & add\_4{[}0{]}{[}0{]}                                             \\
			activation\_6 (Activation)                    & multiple                    & 0                 & conv2d\_19{[}0{]}{[}0{]}                                         \\
			batch\_normalization\_17 (BatchNormalization) & multiple                    & 256               & activation\_6{[}0{]}{[}0{]}                                      \\
			conv2d\_20 (Conv2D)                           & multiple                    & 36928             & batch\_normalization\_17{[}0{]}{[}0{]}                           \\
			batch\_normalization\_18 (BatchNormalization) & multiple                    & 256               & conv2d\_20{[}0{]}{[}0{]}                                         \\
			add\_5 (Add)                                  & multiple                    & 0                 & batch\_normalization\_18{[}0{]}{[}0{]}, add\_4{[}0{]}{[}0{]}     \\
			conv2d\_21 (Conv2D)                           & multiple                    & 36928             & add\_5{[}0{]}{[}0{]}                                             \\
			activation\_7 (Activation)                    & multiple                    & 0                 & conv2d\_21{[}0{]}{[}0{]}                                         \\
			batch\_normalization\_19 (BatchNormalization) & multiple                    & 256               & activation\_7{[}0{]}{[}0{]}                                      \\
			conv2d\_22 (Conv2D)                           & multiple                    & 36928             & batch\_normalization\_19{[}0{]}{[}0{]}                           \\
			batch\_normalization\_20 (BatchNormalization) & multiple                    & 256               & conv2d\_22{[}0{]}{[}0{]}                                         \\
			add\_6 (Add)                                  & multiple                    & 0                 & batch\_normalization\_20{[}0{]}{[}0{]}, add\_5{[}0{]}{[}0{]}     \\
			conv2d\_23 (Conv2D)                           & multiple                    & 36928             & add\_6{[}0{]}{[}0{]}                                             \\
			activation\_8 (Activation)                    & multiple                    & 0                 & conv2d\_23{[}0{]}{[}0{]}                                         \\
			batch\_normalization\_21 (BatchNormalization) & multiple                    & 256               & activation\_8{[}0{]}{[}0{]}                                      \\
			conv2d\_24 (Conv2D)                           & multiple                    & 36928             & batch\_normalization\_21{[}0{]}{[}0{]}                           \\
			batch\_normalization\_22 (BatchNormalization) & multiple                    & 256               & conv2d\_24{[}0{]}{[}0{]}                                         \\
			\\ \bottomrule
		\end{tabular}
	}
\end{table}

\begin{table}
	\centering

	{\tiny
		\begin{tabular}{@{}llll@{}}
			\toprule
			\textbf{Layer (type)}                         & \textbf{Output Shape}       & \textbf{Param \#} & \textbf{Connected to}                                            \\ \midrule

			add\_7 (Add)                                  & multiple                    & 0                 & batch\_normalization\_22{[}0{]}{[}0{]}, add\_6{[}0{]}{[}0{]}     \\
			conv2d\_25 (Conv2D)                           & multiple                    & 36928             & add\_7{[}0{]}{[}0{]}                                             \\
			activation\_9 (Activation)                    & multiple                    & 0                 & conv2d\_25{[}0{]}{[}0{]}                                         \\
			batch\_normalization\_23 (BatchNormalization) & multiple                    & 256               & activation\_9{[}0{]}{[}0{]}                                      \\
			conv2d\_26 (Conv2D)                           & multiple                    & 36928             & batch\_normalization\_23{[}0{]}{[}0{]}                           \\
			batch\_normalization\_24 (BatchNormalization) & multiple                    & 256               & conv2d\_26{[}0{]}{[}0{]}                                         \\
			add\_8 (Add)                                  & multiple                    & 0                 & batch\_normalization\_24{[}0{]}{[}0{]}, add\_7{[}0{]}{[}0{]}     \\
			conv2d\_27 (Conv2D)                           & multiple                    & 36928             & add\_8{[}0{]}{[}0{]}                                             \\
			activation\_10 (Activation)                   & multiple                    & 0                 & conv2d\_27{[}0{]}{[}0{]}                                         \\
			batch\_normalization\_25 (BatchNormalization) & multiple                    & 256               & activation\_10{[}0{]}{[}0{]}                                     \\
			conv2d\_28 (Conv2D)                           & multiple                    & 36928             & batch\_normalization\_25{[}0{]}{[}0{]}                           \\
			batch\_normalization\_26 (BatchNormalization) & multiple                    & 256               & conv2d\_28{[}0{]}{[}0{]}                                         \\
			add\_9 (Add)                                  & multiple                    & 0                 & batch\_normalization\_26{[}0{]}{[}0{]}, add\_8{[}0{]}{[}0{]}     \\
			conv2d\_29 (Conv2D)                           & multiple                    & 36928             & add\_9{[}0{]}{[}0{]}                                             \\
			activation\_11 (Activation)                   & multiple                    & 0                 & conv2d\_29{[}0{]}{[}0{]}                                         \\
			batch\_normalization\_27 (BatchNormalization) & multiple                    & 256               & activation\_11{[}0{]}{[}0{]}                                     \\
			conv2d\_30 (Conv2D)                           & multiple                    & 36928             & batch\_normalization\_27{[}0{]}{[}0{]}                           \\
			batch\_normalization\_28 (BatchNormalization) & multiple                    & 256               & conv2d\_30{[}0{]}{[}0{]}                                         \\
			add\_10 (Add)                                 & multiple                    & 0                 & batch\_normalization\_28{[}0{]}{[}0{]}, add\_9{[}0{]}{[}0{]}     \\
			conv2d\_31 (Conv2D)                           & multiple                    & 36928             & add\_10{[}0{]}{[}0{]}                                            \\
			activation\_12 (Activation)                   & multiple                    & 0                 & conv2d\_31{[}0{]}{[}0{]}                                         \\
			batch\_normalization\_29 (BatchNormalization) & multiple                    & 256               & activation\_12{[}0{]}{[}0{]}                                     \\
			conv2d\_32 (Conv2D)                           & multiple                    & 36928             & batch\_normalization\_29{[}0{]}{[}0{]}                           \\
			batch\_normalization\_30 (BatchNormalization) & multiple                    & 256               & conv2d\_32{[}0{]}{[}0{]}                                         \\
			add\_11 (Add)                                 & multiple                    & 0                 & batch\_normalization\_30{[}0{]}{[}0{]}, add\_10{[}0{]}{[}0{]}    \\
			conv2d\_33 (Conv2D)                           & multiple                    & 36928             & add\_11{[}0{]}{[}0{]}                                            \\
			activation\_13 (Activation)                   & multiple                    & 0                 & conv2d\_33{[}0{]}{[}0{]}                                         \\
			batch\_normalization\_31 (BatchNormalization) & multiple                    & 256               & activation\_13{[}0{]}{[}0{]}                                     \\
			conv2d\_34 (Conv2D)                           & multiple                    & 36928             & batch\_normalization\_31{[}0{]}{[}0{]}                           \\
			batch\_normalization\_32 (BatchNormalization) & multiple                    & 256               & conv2d\_34{[}0{]}{[}0{]}                                         \\
			add\_12 (Add)                                 & multiple                    & 0                 & batch\_normalization\_32{[}0{]}{[}0{]}, add\_11{[}0{]}{[}0{]}    \\
			conv2d\_35 (Conv2D)                           & multiple                    & 36928             & add\_12{[}0{]}{[}0{]}                                            \\
			activation\_14 (Activation)                   & multiple                    & 0                 & conv2d\_35{[}0{]}{[}0{]}                                         \\
			batch\_normalization\_33 (BatchNormalization) & multiple                    & 256               & activation\_14{[}0{]}{[}0{]}                                     \\
			conv2d\_36 (Conv2D)                           & multiple                    & 36928             & batch\_normalization\_33{[}0{]}{[}0{]}                           \\
			batch\_normalization\_34 (BatchNormalization) & multiple                    & 256               & conv2d\_36{[}0{]}{[}0{]}                                         \\
			add\_13 (Add)                                 & multiple                    & 0                 & batch\_normalization\_34{[}0{]}{[}0{]}, add\_12{[}0{]}{[}0{]}    \\
			conv2d\_37 (Conv2D)                           & multiple                    & 36928             & add\_13{[}0{]}{[}0{]}                                            \\
			activation\_15 (Activation)                   & multiple                    & 0                 & conv2d\_37{[}0{]}{[}0{]}                                         \\
			batch\_normalization\_35 (BatchNormalization) & multiple                    & 256               & activation\_15{[}0{]}{[}0{]}                                     \\
			conv2d\_38 (Conv2D)                           & multiple                    & 36928             & batch\_normalization\_35{[}0{]}{[}0{]}                           \\
			batch\_normalization\_36 (BatchNormalization) & multiple                    & 256               & conv2d\_38{[}0{]}{[}0{]}                                         \\
			add\_14 (Add)                                 & multiple                    & 0                 & batch\_normalization\_36{[}0{]}{[}0{]}, add\_13{[}0{]}{[}0{]}    \\
			conv2d\_39 (Conv2D)                           & multiple                    & 36928             & add\_14{[}0{]}{[}0{]}                                            \\
			activation\_16 (Activation)                   & multiple                    & 0                 & conv2d\_39{[}0{]}{[}0{]}                                         \\
			batch\_normalization\_37 (BatchNormalization) & multiple                    & 256               & activation\_16{[}0{]}{[}0{]}                                     \\
			conv2d\_40 (Conv2D)                           & multiple                    & 36928             & batch\_normalization\_37{[}0{]}{[}0{]}                           \\
			batch\_normalization\_38 (BatchNormalization) & multiple                    & 256               & conv2d\_40{[}0{]}{[}0{]}                                         \\
			add\_15 (Add)                                 & multiple                    & 0                 & batch\_normalization\_38{[}0{]}{[}0{]}, add\_14{[}0{]}{[}0{]}    \\
			conv2d\_41 (Conv2D)                           & multiple                    & 36928             & add\_15{[}0{]}{[}0{]}                                            \\
			batch\_normalization\_39 (BatchNormalization) & multiple                    & 256               & conv2d\_41{[}0{]}{[}0{]}                                         \\
			add\_16 (Add)                                 & multiple                    & 0                 & batch\_normalization\_39{[}0{]}{[}0{]}, activation{[}0{]}{[}0{]} \\
			conv2d\_42 (Conv2D)                           & multiple                    & 147712            & add\_16{[}0{]}{[}0{]}                                            \\
			activation\_17 (Activation)                   & multiple                    & 0                 & conv2d\_42{[}0{]}{[}0{]}                                         \\
			conv2d\_43 (Conv2D)                           & multiple                    & 62211             & activation\_17{[}0{]}{[}0{]}                                     \\ \midrule
			Total params                                  &                             & 1,452,611         &                                                                  \\
			Trainable params                              &                             & 1,448,387         &                                                                  \\
			Non-trainable params                          &                             & 4,224             &                                                                  \\ \bottomrule
		\end{tabular}
	}
\end{table}
\clearpage
\subsection{TLGAN discriminator network architectures}\label{sup:disnetworkarchitecture}
% \usepackage{booktabs}
\begin{table}[htp]
	\centering
	{\small
		\begin{tabular}{@{}lll@{}}
			\toprule
			\textbf{Layer (type)}                        & \textbf{Output Shape}   & \textbf{Param \#} \\ \midrule
			input\_3 (InputLayer)                        & {[}(None, 64, 64, 3){]} & 0                 \\
			conv2d (Conv2D)                              & (None, 64, 64, 64)      & 1792              \\
			leaky\_re\_lu (LeakyReLU)                    & (None, 64, 64, 64)      & 0                 \\
			conv2d\_1 (Conv2D)                           & (None, 32, 32, 64)      & 36928             \\
			leaky\_re\_lu\_1 (LeakyReLU)                 & (None, 32, 32, 64)      & 0                 \\
			batch\_normalization (BatchNormalization)    & (None, 32, 32, 64)      & 256               \\
			conv2d\_2 (Conv2D)                           & (None, 32, 32, 128)     & 73856             \\
			leaky\_re\_lu\_2 (LeakyReLU)                 & (None, 32, 32, 128)     & 0                 \\
			batch\_normalization\_1 (BatchNormalization) & (None, 32, 32, 128)     & 512               \\
			conv2d\_3 (Conv2D)                           & (None, 16, 16, 128)     & 147584            \\
			leaky\_re\_lu\_3 (LeakyReLU)                 & (None, 16, 16, 128)     & 0                 \\
			batch\_normalization\_2 (BatchNormalization) & (None, 16, 16, 128)     & 512               \\
			conv2d\_4 (Conv2D)                           & (None, 16, 16, 256)     & 295168            \\
			leaky\_re\_lu\_4 (LeakyReLU)                 & (None, 16, 16, 256)     & 0                 \\
			batch\_normalization\_3 (BatchNormalization) & (None, 16, 16, 256)     & 1024              \\
			conv2d\_5 (Conv2D)                           & (None, 8, 8, 256)       & 590080            \\
			leaky\_re\_lu\_5 (LeakyReLU)                 & (None, 8, 8, 256)       & 0                 \\
			batch\_normalization\_4 (BatchNormalization) & (None, 8, 8, 256)       & 1024              \\
			conv2d\_6 (Conv2D)                           & (None, 8, 8, 512)       & 1180160           \\
			leaky\_re\_lu\_6 (LeakyReLU)                 & (None, 8, 8, 512)       & 0                 \\
			batch\_normalization\_5 (BatchNormalization) & (None, 8, 8, 512)       & 2048              \\
			conv2d\_7 (Conv2D)                           & (None, 4, 4, 512)       & 2359808           \\
			leaky\_re\_lu\_7 (LeakyReLU)                 & (None, 4, 4, 512)       & 0                 \\
			batch\_normalization\_6 (BatchNormalization) & (None, 4, 4, 512)       & 2048              \\
			dense (Dense)                                & (None, 4, 4, 1024)      & 525312            \\
			leaky\_re\_lu\_8 (LeakyReLU)                 & (None, 4, 4, 1024)      & 0                 \\
			dense\_1 (Dense)                             & (None, 4, 4, 1)         & 1025              \\
			Total params                                 &                         & 5,219,137         \\
			Trainable params                             &                         & 0                 \\
			Non-trainable params                         &                         & 5,219,137         \\ \bottomrule
		\end{tabular}
	}
\end{table}
\clearpage

\subsection{TLGAN feature extraction network architectures (VGG19)}\label{sup:fenetworkarchitecture}
% \usepackage{booktabs}
\begin{table}[htp]
	\centering
	{\small
		\begin{tabular}{@{}lll@{}}
			\toprule
			\textbf{Layer (type)}       & \textbf{Output Shape}   & \textbf{Param \#} \\ \midrule
			input\_1 (InputLayer)       & {[}(None, 64, 64, 3){]} & 0                 \\
			block1\_conv1 (Conv2D)      & (None, 64, 64, 64)      & 1792              \\
			block1\_conv2 (Conv2D)      & (None, 64, 64, 64)      & 36928             \\
			block1\_pool (MaxPooling2D) & (None, 32, 32, 64)      & 0                 \\
			block2\_conv1 (Conv2D)      & (None, 32, 32, 128)     & 73856             \\
			block2\_conv2 (Conv2D)      & (None, 32, 32, 128)     & 147584            \\
			block2\_pool (MaxPooling2D) & (None, 16, 16, 128)     & 0                 \\
			block3\_conv1 (Conv2D)      & (None, 16, 16, 256)     & 295168            \\
			block3\_conv2 (Conv2D)      & (None, 16, 16, 256)     & 590080            \\
			block3\_conv3 (Conv2D)      & (None, 16, 16, 256)     & 590080            \\
			Total params                &                         & 1,735,488         \\
			Trainable params            &                         & 0                 \\
			Non-trainable params        &                         & 1,735,488         \\ \bottomrule
		\end{tabular}
	}
\end{table}
%%%%%%%%%%%%%%%%%%%%%%%%%%%%%%%%%%%%%%%%%%%%%

\end{document}